\documentclass[11pt,letterpaper]{article}
\usepackage{emnlp2016}
\usepackage{times}
\usepackage{latexsym}

\usepackage{url}
\usepackage{tikz}
\usepackage{xcolor}
\usetikzlibrary{arrows}
\usetikzlibrary{decorations.text}
\usepackage{enumitem}

\usepackage{amsmath}
\usepackage{amssymb}
\usepackage{booktabs}
\usepackage{multirow}

\definecolor{nice-red}{HTML}{E41A1C}
\definecolor{nice-orange}{HTML}{FF7F00}
\definecolor{nice-yellow}{HTML}{FFC020}
\definecolor{nice-green}{HTML}{4DAF4A}
\definecolor{nice-blue}{HTML}{377EB8}
\definecolor{nice-purple}{HTML}{984EA3}

\emnlpfinalcopy

\def\emnlppaperid{515}

\title{Stance Detection with Bidirectional Conditional Encoding}

\author{Isabelle Augenstein \and Tim Rockt{\"a}schel\\
	    Department of Computer Science\\
	    University College London\\
	    {\tt i.augenstein@ucl.ac.uk, t.rocktaschel@cs.ucl.ac.uk}\\
	  \AND
	Andreas Vlachos \and Kalina Bontcheva\\
  	Department of Computer Science\\
  	University of Sheffield\\
  {\tt \{a.vlachos, k.bontcheva\}@sheffield.ac.uk}}

\date{}

\begin{document}

\maketitle

\begin{abstract}
Stance detection is the task of classifying the attitude expressed in a text towards a target such as {\em Hillary Clinton} to be ``positive'', ``negative'' or ``neutral''.
Previous work has assumed that either the target is mentioned in the text or that training data for every target is given. This paper considers the more challenging version of this task, where targets are not always mentioned and no training data is available for the test targets.
We experiment with conditional LSTM encoding, which builds a representation of the tweet that is dependent on the target, and demonstrate that it outperforms encoding the tweet and the target independently. Performance is improved further when the conditional model is augmented with bidirectional encoding. We evaluate our approach on the SemEval 2016 Task 6 Twitter Stance Detection corpus achieving performance second best only to a system trained on semi-automatically labelled tweets for the test target. When such weak supervision is added, our approach achieves state--of-the-art results.
\end{abstract}

\section{Introduction}

The goal of stance detection is to classify the attitude expressed in a text towards a given target, as ``positive'', ''negative'', or ''neutral''. 
Such information can be useful for a variety of tasks, e.g.\ \newcite{Mendoza2010} showed that tweets stating actual facts were affirmed by 90\% of the tweets related to them, while tweets conveying false information were predominantly questioned or denied.
In this paper  we focus on a novel stance detection task, namely tweet stance detection towards previously unseen targets (mostly entities such as politicians or issues of public interest), as defined in the SemEval Stance Detection for Twitter task~\cite{StanceSemEval2016}.
This task is rather difficult, firstly due to not having training data for the targets in the test set, and secondly, due to the targets not always being mentioned in the tweet.
For example, the tweet ``@realDonaldTrump is the only honest voice of the @GOP'' expresses a positive stance towards the target {\em Donald Trump}. However, when stance is annotated with respect to {\em Hillary Clinton} as the implicit target, this tweet expresses a negative stance, since supporting candidates from one party implies negative stance towards candidates from other parties.

Thus the challenge is twofold. First, we need to learn a model that interprets the tweet stance towards a target that might not be mentioned in the tweet itself. 
Second, we need to learn such a model without labelled training data for the target with respect to which we are predicting the stance. In the example above, we need to learn a model for {\em Hillary Clinton} by only using training data for other targets. While this renders the task more challenging, it is a more realistic scenario, as it is unlikely that labelled training data for each target of interest will be available.

To address these challenges we develop a neural network architecture based on conditional encoding~\cite{rocktaschel2016reasoning}. A long-short term memory (LSTM) network~\cite{hochreiter1997long} is used to encode the target, followed by a second LSTM that encodes the tweet using the encoding of the target as its initial state. We show that this approach achieves better F1 than an SVM baseline, or an independent LSTM encoding of the tweet and the target.
Results improve further (0.4901 F1) with a bidirectional version of our model, which takes into account the context on either side of the word being encoded. In the context of the shared task, this would have been the second best result, except for an approach which uses automatically labelled tweets for the test targets (F1 of 0.5628). Lastly, when our bidirectional conditional encoding model is trained on such data, it achieves state-of-the-art performance (0.5803 F1). 
 
\section{Task Setup}\label{sec:TaskSetup}

The SemEval 2016 Stance Detection for Twitter shared task~\cite{StanceSemEval2016} consists of two subtasks, Task A and Task B. In Task A the goal is to detect the stance of tweets towards targets given labelled training data for all test targets ({\em Climate Change is a Real Concern}, {\em Feminist Movement}, {\em Atheism}, {\em Legalization of Abortion} and {\em Hillary Clinton}). 
In Task B, which is the focus of this paper, the goal is to detect stance with respect to an unseen target, {\em Donald Trump}, for which labeled training/development data is not provided.

Systems need to classify the stance of each tweet as ``positive'' (FAVOR), ``negative'' (AGAINST) or ``neutral'' (NONE) towards the target.
The official metric reported for the shared task is F1 macro-averaged over the classes FAVOR and AGAINST. Although the F1 of NONE is not considered, systems still need to predict it to avoid precision errors for the other two classes.

Even though participants were not allowed to manually label data for the test target {\em Donald Trump}, they were allowed to label data automatically. The two best-performing systems submitted to Task B, pkudblab~\cite{StanceSemEval2016pkudblab} and LitisMind~\cite{StanceSemEval2016MITRE} made use of this, thus changing the task to weakly supervised seen target stance detection, instead of an unseen target task.
Although the goal of this paper is to present stance detection methods for targets for which no training data is available, we show that they can also be used successfully in a weakly supervised framework and outperform the state-of-the-art on the SemEval 2016 Stance Detection for Twitter dataset.

\section{Methods}\label{sec:Methods}

A common stance detection approach is to treat it as a sentence-level classification task similar to sentiment analysis~\cite{pang2008opinion,socher-EtAl:2013:EMNLP}. However, such an approach cannot capture the stance of a tweet with respect to a particular target, unless training data is available for each of the test targets. In such cases, we could learn that a tweet mentioning {\em Donald Trump} in a positive manner expresses a negative stance towards {\em Hillary Clinton}.
Despite this limitation, we use two such baselines, one implemented with a Support Vector Machine (SVM) classifier and one with an LSTM network, in order to assess whether we are successful in incorporating the target in stance prediction.

A naive approach to incorporate the target in stance prediction would be to generate features concatenating the target with words from the tweet.
Ignoring the issue that such features would be rather sparse, a classifier could learn that some words have target-dependent stance weights, but it still assumes that training data is available for each target.

In order to learn how to combine the stance target with the tweet in a way that generalises to unseen targets, we focus on learning distributed representations and ways to combine them.
The following sections develop progressively the proposed bidirectional conditional LSTM encoding  model, starting from independently encoding the tweet and the target using LSTMs.

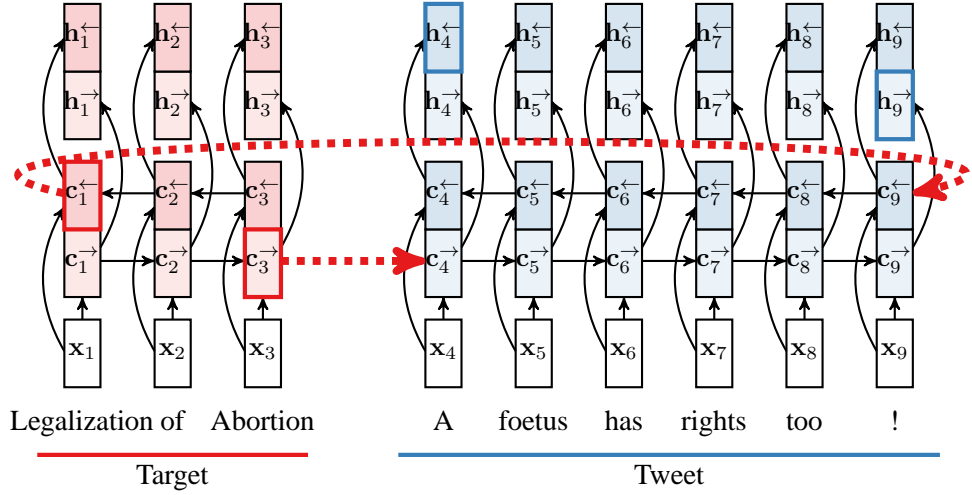
\begin{figure*}
\hspace*{-1.5cm}
\begin{tikzpicture}[scale=1.2]
\foreach \i/\l in {1/1,2/2,3/3} {
  \path[draw, thick] (\i-0.2,0) rectangle (\i+0.2,0.75) {};
  \path[draw, thick, fill=nice-red!10] (\i-0.2,1) rectangle (\i+0.2,1.75) {};
  \path[draw, thick, fill=nice-red!20] (\i-0.2,1.75) rectangle (\i+0.2,2.5) {};  
  \path[draw, thick, fill=nice-red!10] (\i-0.2,2.75) rectangle (\i+0.2,3.5) {};
  \path[draw, thick, fill=nice-red!20] (\i-0.2,3.5) rectangle (\i+0.2,4.25) {};
  \draw[->, >=stealth', thick] (\i,0.75) -- (\i,1);
  \draw[->, >=stealth', thick] (\i-0.2,0.4) to[bend left=30] (\i-0.2,2);
  \draw[->, >=stealth', thick] (\i-0.2,2.3) to[bend left=30] (\i-0.2,3.87);
  \draw[->, >=stealth', thick] (\i+0.2,1.55) to[bend right=30] (\i+0.2,3.17);  
  \node[] at (\i,0.4) {\small$\mathbf{x}_\l$};
  \node[] at (\i,1.4) {\small$\mathbf{c}^\rightarrow_\l$};
  \node[] at (\i,2.15) {\small$\mathbf{c}^\leftarrow_\l$};
  \node[] at (\i,3.15) {\small$\mathbf{h}^\rightarrow_\l$};
  \node[] at (\i,3.9) {\small$\mathbf{h}^\leftarrow_\l$};
}
\foreach \i/\l in {5/4,6/5,7/6,8/7,9/8,10/9} {
  \path[draw, thick] (\i-0.2,0) rectangle (\i+0.2,0.75) {};
  \path[draw, thick, fill=nice-blue!10] (\i-0.2,1) rectangle (\i+0.2,1.75) {};
  \path[draw, thick, fill=nice-blue!20] (\i-0.2,1.75) rectangle (\i+0.2,2.5) {};  
  \path[draw, thick, fill=nice-blue!10] (\i-0.2,2.75) rectangle (\i+0.2,3.5) {};
  \path[draw, thick, fill=nice-blue!20] (\i-0.2,3.5) rectangle (\i+0.2,4.25) {};
  \draw[->, >=stealth', thick] (\i,0.75) -- (\i,1);
  \draw[->, >=stealth', thick] (\i-0.2,0.4) to[bend left=30] (\i-0.2,2);
  \draw[->, >=stealth', thick] (\i-0.2,2.3) to[bend left=30] (\i-0.2,3.87);
  \draw[->, >=stealth', thick] (\i+0.2,1.55) to[bend right=30] (\i+0.2,3.17);  
  \node[] at (\i,0.4) {\small$\mathbf{x}_{\l}$};
  \node[] at (\i,1.4) {\small$\mathbf{c}^\rightarrow_{\l}$};
  \node[] at (\i,2.15) {\small$\mathbf{c}^\leftarrow_{\l}$};
  \node[] at (\i,3.15) {\small$\mathbf{h}^\rightarrow_{\l}$};
  \node[] at (\i,3.9) {\small$\mathbf{h}^\leftarrow_{\l}$};
}
\foreach \i in {1,2,5,6,7,8,9} {
  \draw[->, >=stealth', thick] (\i+0.2,1.4) -- (\i+1-0.2,1.4);
  \draw[->, >=stealth', thick] (\i+1-0.2,2.15) -- (\i+0.2,2.15);
}
\foreach \i/\word in {1/Legalization, 2/of, 3/Abortion, 5/A, 6/foetus, 7/has, 8/rights, 9/too, 10/!} {
  \node[anchor=north, text height=1.5ex, text depth=.25ex, yshift=-2em] at (\i, 0.5) {\word};
  %\node[anchor=north, text height=1.5ex, text depth=.25ex, yshift=-2em] at (\i, {Mod(\i,2)*0.5}) {\word};
  %\draw[->] (\i, {Mod(\i,2)*0.5-0.7}) -- (\i,0);
}

\draw[ultra thick, nice-red] (0.5,-0.75) -- (3.5,-0.75);
\draw[ultra thick, nice-blue] (4.5,-0.75) -- (10.5,-0.75);
\node[anchor=north] at (2,-0.75) {Target};
\node[anchor=north] at (7.5,-0.75) {Tweet};
\draw[->, >=stealth', line width=3pt, color=nice-red, dashed] (3+0.2,1.4) -- (4+1-0.2,1.4);
\draw[->, >=stealth', line width=3pt, color=nice-red, dashed] (1-0.2,2.15) to[bend left=168] (10+0.2,2.15);

\path[draw, ultra thick, color=nice-blue] (5-0.2,3.5) rectangle (5+0.2,4.25) {};
\path[draw, ultra thick, color=nice-blue] (10-0.2,2.75) rectangle (10+0.2,3.5) {};
\path[draw, ultra thick, color=nice-red] (1-0.2,1.75) rectangle (1+0.2,2.5) {};  
\path[draw, ultra thick, color=nice-red] (3-0.2,1) rectangle (3+0.2,1.75) {};
\end{tikzpicture}
\caption{Bidirectional encoding of tweet conditioned on bidirectional encoding of target ($[\mathbf{c}^\rightarrow_3\;\mathbf{c}^\leftarrow_1]$). The stance is predicted using the last forward and reversed output representations ($[\mathbf{h}^\rightarrow_{9}\;\mathbf{h}^\leftarrow_4]$).}
\label{fig:cond}
\end{figure*}

\subsection{Independent Encoding}
\label{sec:concat}

Our initial attempt to learn distributed representations for the tweets and the targets is to encode the target and tweet independently as $k$-dimensional dense vectors using two LSTMs~\cite{hochreiter1997long}.
%LSTMs learn to encode sequences by using memory cells that store information, as well as gates that control the flow of information. 
  \begin{align*}
    \mathbf{H} &= \left[{
      \begin{array}{*{20}c}
        \mathbf{x}_t \\
        \mathbf{h}_{t-1}
      \end{array} }
    \right]\\
    \mathbf{i}_t &= \sigma(\mathbf{W}^i\mathbf{H}+\mathbf{b}^i)\\
    \mathbf{f}_t &= \sigma(\mathbf{W}^f\mathbf{H}+\mathbf{b}^f)\\
    \mathbf{o}_t &= \sigma(\mathbf{W}^o\mathbf{H}+\mathbf{b}^o)\\
    \mathbf{c}_t &= \mathbf{f}_t \odot \mathbf{c}_{t-1} + \mathbf{i}_t \odot
    \tanh(\mathbf{W}^c\mathbf{H}+\mathbf{b}^c)\\
    \mathbf{h}_t &= \mathbf{o}_t \odot \tanh(\mathbf{c}_t)
  \end{align*}
Here, $\mathbf{x}_t$ is an input vector at time step $t$, $\mathbf{c}_t$ denotes the LSTM memory, $\mathbf{h}_t \in \mathbb{R}^k$ is an output vector and the remaining weight matrices and biases are trainable parameters.
We concatenate the two output vector representations and classify the stance using the softmax over a non-linear projection 
\[
\text{softmax}(\text{tanh}(\mathbf{W}^\text{ta}\mathbf{h}_\text{target}+\mathbf{W}^\text{tw}\mathbf{h}_\text{tweet} + \mathbf{b}))
\] 
into the space of the three classes for stance detection where $\mathbf{W}^\text{ta}, \mathbf{W}^\text{tw} \in \mathbb{R}^{3\times k}$ are trainable weight matrices and $\mathbf{b}\in\mathbb{R}^3$ is a trainable class bias. 
This model learns target-independent distributed representations for the tweets and relies on the non-linear projection layer to incorporate the target in the stance prediction.

\subsection{Conditional Encoding}\label{sec:cond_enc_LSTM}
In order to learn target-dependent tweet representations, we use conditional encoding as previously applied to the task of recognising textual entailment \cite{rocktaschel2016reasoning}. We use one LSTM to encode the target as a fixed-length vector. Then, we encode the tweet with another LSTM, whose state is initialised with the representation of the target. Finally, we use the last output vector of the tweet LSTM to predict the stance of the target-tweet pair. 

 Formally, let $(\mathbf{x}_1, \ldots, \mathbf{x}_T)$ be a sequence of target word vectors, $(\mathbf{x}_{T+1}, \ldots, \mathbf{x}_N)$ be a sequence of tweet word vectors and $[\mathbf{h}_0\;\mathbf{c}_0]$ be a start state of zeros. The two LSTMs map input vectors and a previous state to a next state as follows:
 \begin{align*}
 	[\mathbf{h}_1\;\mathbf{c}_1] &= \text{LSTM}^\text{target}(\mathbf{x}_1, \mathbf{h}_0, \mathbf{c}_0)\\
 	&\ldots\\
 [\mathbf{h}_T\;\mathbf{c}_T] &= \text{LSTM}^\text{target}(\mathbf{x}_T, \mathbf{h}_{T-1}, \mathbf{c}_{T-1})\\
      [\mathbf{h}_{T+1}\;\mathbf{c}_{T+1}] &= \text{LSTM}^\text{tweet}(\mathbf{x}_{T+1}, \mathbf{h}_0, \mathbf{c}_T)\\
     &\ldots\\
      [\mathbf{h}_N\;\mathbf{c}_N] &= \text{LSTM}^\text{tweet}(\mathbf{x}_{N}, \mathbf{h}_{N-1}, \mathbf{c}_{N-1})
 \end{align*}
 Finally, the stance of the tweet w.r.t. the target is classified using a non-linear projection
 \begin{align*}
 \mathbf{c} &= \text{tanh}(\mathbf{W} \mathbf{h}_N)
 \end{align*}
 where $\mathbf{W}\in\mathbb{R}^{3\times k}$ is a trainable weight matrix.
This effectively allows the second LSTM to read the tweet in a target-specific manner, which is crucial since the stance of the tweet depends on the target (recall the Donald Trump example above). 

\subsection{Bidirectional Conditional Encoding}
Bidirectional LSTMs \cite{graves2005framewise} have been shown to learn improved representations of sequences by encoding a sequence from left to right and from right to left. Therefore, we adapt the conditional encoding model from Section~\ref{sec:cond_enc_LSTM} to use bidirectional LSTMs, which represent the target and the tweet using two vectors for each of them, one obtained by reading the target and then the tweet left-to-right (as in the conditional LSTM encoding) and one obtained by reading them right-to-left. To achieve this, we initialise the state of the bidirectional LSTM that reads the tweet by the last state of the forward and reversed encoding of the target  (see Figure~\ref{fig:cond}). 
The bidirectional encoding allows the model to construct target-dependent representations of the tweet such that when a word is considered, both its left- and the right-hand side context are taken into account.

\subsection{Unsupervised Pretraining}\label{sec:Pretraining}

In order to counter-balance the relatively small amount of training data available (5,628 instances in total), we employ unsupervised pre-training by initialising the word embeddings used in the LSTMs with an appropriately trained word2vec model \cite{mikolov2013efficient}.  Note that these embeddings are used only for initialisation, as we allow them to be optimised further during training.

In more detail, we train a word2vec model on a corpus of 395,212 unlabelled tweets, collected with the Twitter Keyword Search API\footnote{\url{https://dev.twitter.com/rest/public/} \\ \-\hspace{.75cm} \url{search}} between November 2015 and January 2016, plus all the tweets contained in the official SemEval 2016 Stance Detection datasets~\cite{StanceSemEval2016}.
 The unlabelled tweets are collected so that they contain the targets considered in the shared task, using up to two keywords per target, namely ``hillary'', ``clinton'', ``trump'', ``climate'', ``femini'', ``aborti''.
Note that Twitter does not allow for regular expression search, so this is a free text search disregarding possible word boundaries.
We combine this large unlabelled corpus with the official training data and train a skip-gram word2vec model (dimensionality 100, 5 min words, context window of 5).

Tweets and targets are tokenised with the Twitter-adapted tokeniser twokenize\footnote{\url{https://github.com/leondz/twokenize}}. Subsequently, all tokens are  lowercased, URLs are removed, and stopword tokens are filtered (i.e. punctuation characters, Twitter-specific stopwords (``rt'', ``\#semst'', ``via'').

As it will be shown in our experiments, unsupervised pre-training is quite helpful, since it is difficult to learn representations for all the words using only the relatively small training datasets available.

Finally, to ensure that the proposed neural network architectures contribute to the performance, we also use the word vectors from word2vec to develop a Bag-of-Word-Vectors baseline (\textsf{BOWV}), in which the tweet and target representations are fed into a logistic regression classifier with L2 regularization~\cite{scikit-learn}.

\setlength{\tabcolsep}{0.3em}
\begin{table}[t]
\fontsize{10}{10}\selectfont
\begin{center}
\begin{tabular}{l c c c c}
\toprule
\bf Corpus & \bf Favor & \bf Against & \bf None & \bf All  \\ 
\midrule
TaskA\_Tr+Dv & 1462 & 2684 & 1482 & 5628 \\
TaskA\_Tr+Dv\_HC & 224 & 722 & 332 & 1278 \\
TaskB\_Unlab & - & - & - & 278,013 \\
TaskB\_Auto-lab* & 4681 & 5095 & 4026 & 13,802 \\
TaskB\_Test & 148 & 299 & 260 & 707 \\
Crawled\_Unlab* & - & - & - & 395,212 \\
\bottomrule
\end{tabular}
\end{center}
\caption{\label{tab:DataStats} Data sizes of available corpora. \textsf{TaskA\_Tr+Dv\_HC} is the part of \textsf{TaskA\_Tr+Dv} with tweets for the target Hillary Clinton only, which we use for development. \textsf{TaskB\_Auto-lab} is an automatically labelled version of \textsf{TaskB\_Unlab}. Crawled\_Unlab is an unlabelled tweet corpus collected by us.}
\end{table}

\section{Experiments}

Experiments are performed on the SemEval 2016 Task 6 corpus for Stance Detection on Twitter~\cite{StanceSemEval2016}.
We report experiments for two different experimental setups: one is the {\em unseen target} setup (Section~\ref{sec:UnseenTarget}), which is the main focus of this paper, i.e. detecting the stance of tweets towards previously unseen targets. 
We show that conditional encoding, by reading the tweets in a target-specific way, generalises to unseen targets better than baselines which ignore the target.
Next, we compare our approach to previous work in a {\em weakly supervised framework} (Section~\ref{sec:WeaklySup}) and show that our approach outperforms the state-of-the-art on the SemEval 2016 Stance Detection Subtask B corpus.

Table~\ref{tab:DataStats} lists the various corpora used in the experiments and their sizes.
\textsf{TaskA\_Tr+Dv} is the official SemEval 2016 Twitter Stance Detection TaskA training and development corpus, which contain instances for the targets {\em Legalization of Abortion}, {\em Atheism}, {\em Feminist Movement}, {\em Climate Change is a Real Concern} and {\em Hillary Clinton}.
\textsf{TaskA\_Tr+Dv\_HC} is the part of the corpus which contains only the {\em Hillary Clinton} tweets, which we use for development purposes. \textsf{TaskB\_Test} is the TaskB test corpus on which we report results containing {\em Donald Trump} testing instances. \textsf{TaskB\_Unlab} is an unlabelled corpus containing {\em Donald Trump} tweets supplied by the task organisers, and \textsf{TaskB\_Auto-lab*} is an automatically labelled version of a small portion of the corpus for the weakly supervised stance detection experiments reported in Section~\ref{sec:WeaklySup}. Finally, \textsf{Crawled\_Unlab*} is a corpus we collected for unsupervised pre-training (see Section~\ref{sec:Pretraining}).

For all experiments, the official task evaluation script is used.
Predictions are post processed so that if the target is contained in a tweet, the highest-scoring non-neutral stance is chosen.
This was motivated by the observation that in the training data most target-containing tweets express a stance, with only 16\% of them
being neutral. The code used in our experiments is available from  \url{https://github.com/sheffieldnlp/stance-conditional}.

\subsection{Methods}

We compare the following baseline methods: % that do perform conditional encoding:
\begin{itemize}[noitemsep]

\item{SVM trained with word and character tweet n-grams features (\textsf{SVM-ngrams-comb})~\newcite{StanceSemEval2016}}
\item{a majority class baseline (\textsf{Majority baseline}), reported in~\cite{StanceSemEval2016}}
\item{bag of word vectors (\textsf{BoWV}) (see Section~\ref{sec:Pretraining})}
\item{independent encoding of tweet and the target with two LSTMs (\textsf{Concat}) (see Section~\ref{sec:concat})}
\item{encoding of the tweet only with an LSTM (\textsf{TweetOnly}) (see Section~\ref{sec:concat})}
\end{itemize}
to three versions of conditional encoding:
\begin{itemize}[noitemsep]
\item{target conditioned on tweet (\textsf{TarCondTweet})}
\item{tweet conditioned on target (\textsf{TweetCondTar})}
\item{a bidirectional encoding model (\textsf{BiCond})}
\end{itemize}

\setlength{\tabcolsep}{0.4em}
\begin{table}[t]
\fontsize{8}{10}\selectfont
\begin{center}
\begin{tabular}{l c c c c}
\toprule
\bf Method & \bf Stance & \bf P & \bf R & \bf F1 \\
\midrule
\multirow{3}{*}{\textsf{BoWV}} & FAVOR & 0.2444 & 0.0940 & 0.1358 \\
 & AGAINST & 0.5916 & 0.8626 & 0.7019 \\ \cline{2-5}
 & Macro & & & 0.4188 \\
 \midrule
\multirow{3}{*}{\textsf{TweetOnly}} & FAVOR & 0.2127 & 0.5726 & 0.3102 \\
 & AGAINST & 0.6529 & 0.4020 & 0.4976 \\ \cline{2-5}
 & Macro & & & 0.4039 \\
 \midrule
 \multirow{3}{*}{\textsf{Concat}} & FAVOR & 0.1811 & 0.6239 & 0.2808 \\
 & AGAINST & 0.6299 & 0.4504 & 0.5252 \\ \cline{2-5}
 & Macro & & & 0.4030 \\
 \midrule \midrule
 \multirow{3}{*}{\textsf{TarCondTweet}} & FAVOR & 0.3293 & 0.3649 & 0.3462 \\
 & AGAINST & 0.4304 & 0.5686 & 0.4899 \\ \cline{2-5}
 & Macro & & & 0.4180 \\
 \midrule
 \multirow{3}{*}{\textsf{TweetCondTar}} & FAVOR & 0.1985 & 0.2308 & 0.2134 \\
 & AGAINST & 0.6332 & 0.7379 & 0.6816 \\ \cline{2-5} %cmidrule{2-5}
 & Macro & & & 0.4475 \\
 \midrule
  \multirow{3}{*}{\textsf{BiCond}} & FAVOR & 0.2588 & 0.3761 & 0.3066 \\
 & AGAINST & 0.7081 & 0.5802 & 0.6378 \\ \cline{2-5}
& Macro & & & \bf{0.4722} \\
\bottomrule
\end{tabular}
\end{center}
\caption{\label{tab:ResultsDev} Results for the {\em unseen target} stance detection  development setup.}
\end{table}

\setlength{\tabcolsep}{0.4em}
\begin{table}[t]
\fontsize{8}{10}\selectfont
\begin{center}
\begin{tabular}{l c c c c}
\toprule
\bf Method & \bf Stance & \bf P & \bf R & \bf F1 \\
\midrule
\multirow{3}{*}{\textsf{BoWV}} & FAVOR & 0.3158 & 0.0405 & 0.0719 \\
 & AGAINST & 0.4316 & 0.8963 & 0.5826 \\ \cline{2-5}
 & Macro & & & 0.3272 \\
 \midrule
 \multirow{3}{*}{\textsf{TweetOnly}} & FAVOR & 0.2767 & 0.3851 & 0.3220 \\
 & AGAINST & 0.4225 & 0.5284 & 0.4695 \\ \cline{2-5}
 & Macro & & & 0.3958 \\
  \midrule
 \multirow{3}{*}{\textsf{Concat}} & FAVOR & 0.3145 & 0.5270 & 0.3939 \\
 & AGAINST & 0.4452 & 0.4348 & 0.4399 \\ \cline{2-5}
 & Macro & & & 0.4169 \\
 \midrule \midrule
\multirow{3}{*}{\textsf{TarCondTweet}} & FAVOR & 0.2322 & 0.4188 & 0.2988 \\
 & AGAINST & 0.6712 & 0.6234 & 0.6464 \\ \cline{2-5}
 & Macro & & & 0.4726 \\
\midrule
\multirow{3}{*}{\textsf{TweetCondTar}} & FAVOR & 0.3710 & 0.5541 & 0.4444 \\
 & AGAINST & 0.4633 & 0.5485 & 0.5023 \\ \cline{2-5}
& Macro & & & 0.4734 \\
\midrule
\multirow{3}{*}{\textsf{BiCond}} & FAVOR & 0.3033 & 0.5470 & 0.3902 \\
 & AGAINST & 0.6788 & 0.5216 & 0.5899 \\ \cline{2-5}
 & Macro & & & \bf{0.4901} \\
\bottomrule
\end{tabular}
\end{center}
\caption{\label{tab:ResultsTest} Results for the {\em unseen target} stance detection  test setup.}
\end{table}

\section{Unseen Target Stance Detection}\label{sec:UnseenTarget}

As explained earlier, the challenge is to learn a model without any manually labelled training data for the test target, but only using the data from the Task A targets. In order to avoid using any labelled data for {\em Donald Trump}, while still having a (labelled) development set to tune and evaluate our models, we used the tweets labelled for {\em Hillary Clinton} as a development set and the tweets for the remaining four targets as training. We refer to this as the \emph{development setup}, and all models are tuned using this setup.
The labelled {\em Donald Trump} tweets were only used in reporting our final results.

For the final results we train on all the data from the development setup and evaluate on the official Task B test set, i.e. the {\em Donald Trump} tweets. We refer to this as our \emph{test setup}.

Based on a small grid search using the development setup, the following settings for LSTM-based models were chosen: 
input layer size 100 (equal to the word embedding dimension), hidden layer size of 60, training for max 50 epochs with initial learning rate 1e-3 using ADAM~\cite{journals/corr/KingmaB14} for optimisation, dropout 0.1. Models were trained using cross-entropy loss.
The use of one, relatively small hidden layer and dropout help to avoid overfitting.

\subsection{Results and Discussion}

\setlength{\tabcolsep}{0.4em}
\begin{table}[!t]
\fontsize{8}{10}\selectfont
\begin{center}
\begin{tabular}{l l c c c c}
\toprule
\bf EmbIni & \bf NumMatr & \bf Stance & \bf P & \bf R & \bf F1 \\
\midrule
\multirow{6}{*}{\textsf{Random}} & \multirow{3}{*}{\textsf{Sing}} & FAVOR & 0.1982 & 0.3846 & 0.2616 \\
 & & AGAINST & 0.6263 & 0.5929 & 0.6092 \\ \cline{3-6}
 & & Macro & & & 0.4354 \\
\cmidrule{2-6}
 & \multirow{3}{*}{\textsf{Sep}} & FAVOR & 0.2278 & 0.5043 & 0.3138 \\
& & AGAINST & 0.6706 & 0.4300 & 0.5240 \\ \cline{3-6}
& & Macro & & & 0.4189 \\
\midrule
\multirow{6}{*}{\textsf{PreFixed}} & \multirow{3}{*}{\textsf{Sing}} & FAVOR & 0.6000 & 0.0513 & 0.0945 \\
& & AGAINST & 0.5761 & 0.9440 & 0.7155 \\ \cline{3-6}
& & Macro & & & 0.4050 \\
\cmidrule{2-6}
& \multirow{3}{*}{\textsf{Sep}} & FAVOR & 0.1429 & 0.0342 & 0.0552 \\
& & AGAINST & 0.5707 & 0.9033 & 0.6995 \\ \cline{3-6}
& & Macro & & & 0.3773 \\
\midrule
\multirow{6}{*}{\textsf{PreCont}} & \multirow{3}{*}{\textsf{Sing}} & FAVOR & 0.2588 & 0.3761 & 0.3066 \\
& & AGAINST & 0.7081 & 0.5802 & 0.6378 \\ \cline{3-6}
& & Macro & & & \bf{0.4722} \\
\cmidrule{2-6}
& \multirow{3}{*}{\textsf{Sep}} & FAVOR & 0.2243 & 0.4103 & 0.2900 \\
& & AGAINST & 0.6185 & 0.5445 & 0.5792 \\ \cline{3-6}
& & Macro & & & 0.4346 \\
\bottomrule
\end{tabular}
\end{center}
\caption{\label{tab:ResultsEmbIni} Results for the {\em unseen target} stance detection development setup using \textsf{BiCond}, with single vs separate embeddings matrices for tweet and target and different initialisations}
\end{table}

Results for the unseen target setting show how well conditional encoding is suited for learning target-dependent representations of tweets, and crucially, how well such representations generalise to unseen targets.
The best performing method on both development (Table~\ref{tab:ResultsDev}) and test setups (Table~\ref{tab:ResultsTest}) is \textsf{BiCond}, which achieves an F1 of 0.4722 and 0.4901 respectively. Notably, \textsf{Concat}, which learns an independent encoding of the target and the tweets, does not achieve big F1 improvements over \textsf{TweetOnly}, which learns a representation of the tweets only. This shows that it is not sufficient to just take the target into account, but is is important to learn target-dependent encodings for the tweets. 
Models that learn to condition the encoding of tweets on targets outperform all baselines on the test set. 

It is further worth noting that the Bag-of-Word-Vectors baseline achieves results comparable with \textsf{TweetOnly}, \textsf{Concat} and one of the conditional encoding models, \textsf{TarCondTweet}, on the dev set, even though it achieves significantly lower performance on the test set. This indicates that the pre-trained word embeddings on their own are already very useful for stance detection. This is consistent with findings of other works showing the usefulness of such a Bag-of-Word-Vectors baseline for the related tasks of recognising textual entailment \newcite{bowman-EtAl:2015:EMNLP} and sentiment analysis \newcite{Eisner2016Emoji}.

Our best result in the test setup with \textsf{BiCond} is the second highest reported result on the Twitter Stance Detection corpus, however the first, third and fourth best approaches achieved their results by automatically labelling {\em Donald Trump} training data.  \textsf{BiCond} for the unseen target setting outperforms the third and fourth best approaches by a large margin (5 and 7 points in Macro F1, respectively), as can be seen in Table~\ref{tab:ResultsTestSoA}. Results for weakly supervised stance detection are discussed in Section~\ref{sec:weak}.

\setlength{\tabcolsep}{0.35em}
\begin{table}[!t]
\fontsize{8}{10}\selectfont
\begin{center}
\begin{tabular}{l l c c c c}
\toprule
\bf Method & \bf inTwe & \bf Stance & \bf P & \bf R & \bf F1 \\
\midrule
 \multirow{6}{*}{\textsf{Concat}} & \multirow{3}{*}{\textsf{Yes}} & FAVOR     & 0.3153 & 0.6214 & 0.4183 \\
& & AGAINST & 0.7438 & 0.4630 & 0.5707 \\ \cline{3-6} %\cmidrule{3-6}
 & & Macro & & & 0.4945 \\
\cmidrule{2-6}
 & \multirow{3}{*}{\textsf{No}} & FAVOR & 0.0450 & 0.6429 & 0.0841 \\
 & & AGAINST & 0.4793 & 0.4265 & 0.4514 \\ \cline{3-6}
& & Macro & & & 0.2677 \\
 \midrule
 \multirow{6}{*}{\textsf{TweetCondTar}} & \multirow{3}{*}{\textsf{Yes}} & FAVOR & 0.3529 & 0.2330 & 0.2807 \\
& & AGAINST & 0.7254 & 0.8327 & 0.7754 \\ \cline{3-6}
& & Macro & & & 0.5280 \\
\cmidrule{2-6}
 & \multirow{3}{*}{\textsf{No}} & FAVOR & 0.0441 & 0.2143 & 0.0732 \\
 & & AGAINST & 0.4663 & 0.5588 & 0.5084 \\ \cline{3-6}
 & & Macro & & & 0.2908 \\
 \midrule
 \multirow{6}{*}{\textsf{BiCond}} & \multirow{3}{*}{\textsf{Yes}} & FAVOR & 0.3585 & 0.3689 & 0.3636
\\
& & AGAINST & 0.7393 & 0.7393 & 0.7393 \\ \cline{3-6}
&  & Macro & & & \bf{0.5515} \\
\cmidrule{2-6}
 & \multirow{3}{*}{\textsf{No}} & FAVOR & 0.0938 & 0.4286 & 0.1538 \\
 & & AGAINST & 0.5846 & 0.2794 & 0.3781 \\ \cline{3-6}
 & & Macro & & & 0.2660 \\
 
\bottomrule
\end{tabular}
\end{center}
\caption{\label{tab:ErrorAna} Results for the {\em unseen target} stance detection development setup for tweets containing the target vs tweets not containing the target.}
\end{table}

\paragraph{Pre-Training}

Table~\ref{tab:ResultsEmbIni} shows the effect of unsupervised pre-training of word embeddings with a word2vec skip-gram model, and furthermore, the results of sharing of these representations between the tweets and targets, on the development set.
The first set of results is with a uniformly \textsf{Random} embedding initialisation in $[-0.1, 0.1]$. \textsf{PreFixed} uses the pre-trained skip-gram word embeddings, whereas \textsf{PreCont} initialises the word embeddings with ones from SkipGram and continues training them during LSTM training. %The best results are achieved with \textsf{PreCont} followed by \textsf{Random}, which means unsupervised pre-training of word embeddings  
Our results show that, in the absence of a large labelled training dataset,  pre-training of word embeddings %on text containing the targets 
is more helpful than random initialisation of embeddings.
\textsf{Sing} vs \textsf{Sep} shows the difference between using shared vs two separate embeddings matrices for looking up the word embeddings. \textsf{Sing} means the word representations for tweet and target vocabularies are shared, whereas \textsf{Sep} means they are different.
Using shared embeddings performs better, which we hypothesise is because the tweets contain some mentions of targets that are tested.

\paragraph{Target in Tweet vs Not in Tweet}

Table~\ref{tab:ErrorAna} shows results on the development set for \textsf{BiCond}, compared to the best unidirectional encoding model, \textsf{TweetCondTar} and the baseline model \textsf{Concat}, split by tweets that contain the target and those that do not.
All three models perform well when the target is mentioned in the tweet, but less so when the targets are not mentioned explicitly. In the case where the target is mentioned in the tweet, biconditional encoding outperforms unidirectional encoding and unidirectional encoding outperforms \textsf{Concat}. This shows that conditional encoding is able to learn useful dependencies between the tweets and the targets.

\setlength{\tabcolsep}{0.4em}
\begin{table}[t]
\fontsize{8}{10}\selectfont
\begin{center}
\begin{tabular}{l c c c c}
\toprule
\bf Method & \bf Stance & \bf P & \bf R & \bf F1 \\
\midrule
\multirow{3}{*}{\textsf{BoWV}} & FAVOR & 0.5156 & 0.6689 & 0.5824 \\
 & AGAINST & 0.6266 & 0.3311 & 0.4333 \\ \cline{2-5}
 & Macro & & & 0.5078 \\
\midrule
\multirow{3}{*}{\textsf{TweetOnly}} & FAVOR & 0.5284 & 0.6284 & 0.5741 \\
& AGAINST & 0.5774 & 0.4615 & 0.5130 \\ \cline{2-5}
& Macro & & & 0.5435 \\
\midrule
\multirow{3}{*}{\textsf{Concat}} & FAVOR & 0.5506 & 0.5878 & 0.5686 \\
 & AGAINST & 0.5794 & 0.4883 & 0.5299 \\ \cline{2-5}
 & Macro & & & 0.5493 \\
 \midrule \midrule
\multirow{3}{*}{\textsf{TarCondTweet}} & FAVOR & 0.5636 & 0.6284 & 0.5942 \\
 & AGAINST & 0.5947 & 0.4515 & 0.5133 \\ \cline{2-5}
& Macro & & & 0.5538 \\
\midrule
\multirow{3}{*}{\textsf{TweetCondTar}} & FAVOR & 0.5868 & 0.6622 & 0.6222 \\
 & AGAINST & 0.5915 & 0.4649 & 0.5206 \\ \cline{2-5}
 & Macro & & & 0.5714 \\
 \midrule
\multirow{3}{*}{\textsf{BiCond}}
& FAVOR & 0.6268 & 0.6014 & 0.6138 \\
& AGAINST & 0.6057 & 0.4983 & 0.5468 \\ \cline{2-5}
& Macro & & & \bf{0.5803} \\
\bottomrule
\end{tabular}
\end{center}
\caption{\label{tab:ResultsTest2} Stance Detection test results for weakly supervised setup, trained on automatically labelled pos+neg+neutral Trump data, and reported on the official test set.}
\end{table}

\section{Weakly Supervised Stance Detection}\label{sec:WeaklySup}
\label{sec:weak}

The previous section showed the usefulness of conditional encoding for unseen target stance detection and compared results against internal baselines. The goal of experiments reported in this section is to compare against participants in the SemEval 2016 Stance Detection Task B. While we consider an {\em unseen target} setup, most submissions, including the three highest ranking ones for Task B, pkudblab~\cite{StanceSemEval2016pkudblab}, LitisMind~\cite{StanceSemEval2016MITRE} and INF-UFRGS~\cite{StanceSemEval2016inf} considered a different experimental setup. They automatically annotated training data for the test target {\em Donald Trump}, thus converting the task into weakly supervised seen target stance detection. The pkudblab system uses a deep convolutional neural network that learns to make 2-way predictions on automatically labelled positive and negative training data for {\em Donald Trump}. The neutral class is predicted according to rules which are applied at test time.

Since the best performing systems which participated in the shared task consider a weakly supervised setup, we further compare our proposed approach to the state-of-the-art using such a weakly supervised setup. Note that, even though pkudblab, LitisMind and INF-UFRGS also use regular expressions to label training data automatically, the resulting datasets were not available to us. Therefore, we had to develop our own automatic labelling method and dataset, which are publicly available from our code repository.

\paragraph{Weakly Supervised Test Setup}
For this setup, the unlabelled {\em Donald Trump} corpus \textsf{TaskB\_Unlab} is annotated automatically. For this purpose we created a small set of regular expressions\footnote{Note that ``$|$'' indiates ``or'', ( ?) indicates optional space}, based on inspection of the \textsf{TaskB\_Unlab} corpus, expressing positive and negative stance towards the target. The regular expressions for the positive stance were:
\begin{itemize}[noitemsep,nolistsep]
\item{make( ?)america( ?)great( ?)again}
\item{trump( ?)(for$|$4)( ?)president}
\item{votetrump}
\item{trumpisright}
\item{the truth}
\item{\#trumprules}
\end{itemize}
The keyphrases for negative stance were: \#dumptrump, \#notrump, \#trumpwatch, racist, idiot, fired\\

A tweet is labelled as positive if one of the positive expressions is detected, else negative if a negative expressions is detected. If neither are detected, the tweet is annotated as neutral randomly with 2\% chance. The resulting corpus size per stance is shown in Table~\ref{tab:DataStats}.
The same hyperparameters for the LSTM-based models are used as for the {\em unseen target} setup described in the previous section.

\subsection{Results and Discussion}

Table~\ref{tab:ResultsTest2} lists our results in the weakly supervised setting.
Table~\ref{tab:ResultsTestSoA} shows all our results, including those using the unseen target setup, compared against the state-of-the-art on the stance detection corpus. Table~\ref{tab:ResultsTestSoA}  further lists baselines reported by~\newcite{StanceSemEval2016}, namely a majority class baseline (\textsf{Majority baseline}), and a method using 1 to 3-gram bag-of-word and character n-gram features (\textsf{SVM-ngrams-comb}), which are extracted from the tweets and used to train a 3-way SVM classifier.

Bag-of-word baselines (\textsf{BoWV}, \textsf{SVM-ngrams-comb}) achieve results comparable to the majority baseline (F1 of 0.2972), which shows how difficult the task is.
The baselines which only extract features from the tweets, \textsf{SVM-ngrams-comb} and \textsf{TweetOnly} perform worse than the baselines which also learn representations for the targets
(\textsf{BoWV}, \textsf{Concat}).
By training conditional encoding models on automatically labelled stance detection data we achieve state-of-the-art results.
The best result (F1 of 0.5803) is achieved with the bi-directional conditional encoding model (\textsf{BiCond}). This shows that such models are suitable for unseen, as well as seen target stance detection. %, followed by automatically labelling positive and negative in-domain training instances and adding those to manually labelled out-of-domain training instances (\textsf{TrumpAdded} setting, best F1 is 0.5699). 
%For \textsf{TrumpAdded}), a bag of word vector baseline achieves surprisingly high results (0.5526).

\setlength{\tabcolsep}{0.4em}
\begin{table}[t]
\fontsize{8}{10}\selectfont
\begin{center}
\begin{tabular}{l c c}
\toprule
\bf Method & \bf Stance & \bf F1 \\
\midrule
\multirow{3}{*}{\textsf{SVM-ngrams-comb} ({\em Unseen Target})} 
 & FAVOR &  0.1842 \\
 & AGAINST & 0.3845 \\  \cline{2-3}
& Macro & 0.2843 \\ 
\midrule
\multirow{3}{*}{\textsf{Majority baseline} ({\em Unseen Target})} 
 & FAVOR &  0.0 \\
 & AGAINST & 0.5944 \\  \cline{2-3}
& Macro & 0.2972 \\ 
\midrule
\multirow{3}{*}{\textsf{BiCond} ({\em{Unseen Target}})} 
& FAVOR & 0.3902 \\
 & AGAINST  & 0.5899 \\  \cline{2-3}
 & Macro & \bf{0.4901} \\ 
 \midrule \midrule
  \multirow{3}{*}{\textsf{INF-UFRGS} ({\em Weakly Supervised*})}
& FAVOR & 0.3256 \\
 & AGAINST  & 0.5209 \\ \cline{2-3}
 & Macro & 0.4232 \\
 \midrule
 \multirow{3}{*}{\textsf{LitisMind} ({\em Weakly Supervised*})}
& FAVOR & 0.3004 \\
 & AGAINST  & 0.5928 \\ \cline{2-3}
 & Macro & 0.4466 \\
 \midrule
 \multirow{3}{*}{\textsf{pkudblab} ({\em Weakly Supervised*})}
& FAVOR & 0.5739 \\
& AGAINST & 0.5517 \\  \cline{2-3}
& Macro & 0.5628 \\ 
\midrule

\multirow{3}{*}{\textsf{BiCond} ({\em Weakly Supervised})}
& FAVOR & 0.6138 \\
& AGAINST & 0.5468 \\  \cline{2-3}
& Macro & \bf{0.5803} \\
\bottomrule
\end{tabular}
\end{center}
\caption{\label{tab:ResultsTestSoA} Stance Detection test results, compared against the state of the art. \textsf{SVM-ngrams-comb} and \textsf{Majority baseline} are reported in~\protect\newcite{StanceSemEval2016}, pkudblab in  \protect\newcite{StanceSemEval2016pkudblab}, LitisMind in  \protect\newcite{StanceSemEval2016MITRE}, INF-UFRGS in  \protect\newcite{StanceSemEval2016inf}}
\end{table}

\section{Related Work}

{\bf Stance Detection}: 
Previous work mostly considered target-specific stance prediction in debates \cite{conf/ijcnlp/HasanN13,walker-EtAl:2012:NAACL-HLT} or student essays~\cite{Faulkner14}.
The task considered in this paper is more challenging than stance detection in debates because, in addition to irregular language, the~\newcite{StanceSemEval2016} dataset is offered without any context, e.g., conversational structure or tweet metadata. The targets are also not always mentioned in the tweets, which is an additional challenge \cite{Augenstein2016SemEval} and distinguishes this task from target-dependent \cite{conf/ijcai/VoZ15,Zhang2016Gated,alghunaim-EtAl:2015:VSM-NLP} and open-domain target-dependent sentiment analysis \cite{mitchell-EtAl:2013:EMNLP,zhang-zhang-vo:2015:EMNLP}. %For the latter, the targets are mentioned, but need to be detected first.
Related work on rumour stance detection either requires training data from the same rumour~\cite{Qazvinian2011}, i.e., target, or is rule-based~\cite{Liu2015} and thus potentially hard to generalise. Finally, the target-dependent stance detection task tackled in this paper is different from that of \newcite{ferreira-vlachos:2016:N16-1}, which while related concerned with the stance of a statement in natural language towards another statement.

{\bf Conditional Encoding}:
Conditional encoding has been applied to the related task of recognising textual entailment~\cite{rocktaschel2016reasoning}, using a dataset of half a million training examples~\cite{bowman-EtAl:2015:EMNLP} and numerous different hypotheses. Our experiments here show that conditional encoding is also successful on a relatively small training set and when applied to an unseen testing target. Moreover, we augment conditional encoding with bidirectional encoding and demonstrate the added benefit of unsupervised pre-training of word embeddings on unlabelled domain data.

\section{Conclusions and Future Work}
This paper showed that conditional LSTM encoding is a successful approach to stance detection for unseen targets. Our unseen target bidirectional conditional encoding approach achieves the second best results reported to date on the SemEval 2016 Twitter Stance Detection corpus. In the weakly supervised seen target scenario, as considered by prior work, our approach achieves the best results to date on the SemEval Task B dataset.
We further show that in the absence of large labelled corpora, unsupervised pre-training can be used to learn target representations for stance detection and improves results on the SemEval corpus. Future work will investigate further the challenge of stance detection for tweets which do not contain explicit mentions of the target.

\section*{Acknowledgments}

This work was partially supported by the European Union under grant agreement No. 611233 {\sc Pheme}\footnote{\texttt{http://www.pheme.eu}} and by Microsoft Research through its PhD Scholarship Programme.

\bibliography{emnlp2016}
\bibliographystyle{emnlp2016}

\end{document}